\documentclass[letterpaper, 10 pt, conference]{ieeeconf}

\IEEEoverridecommandlockouts

\usepackage{amsmath}
\usepackage{amsfonts}
\usepackage{graphicx}
\usepackage{booktabs}
\usepackage{algorithm}
\usepackage{algorithmic}
\usepackage{hyperref}
\usepackage{xcolor}
\usepackage{multirow}
\usepackage{makecell}

\title{\LARGE \bf
Object Reconstruction under Occlusion with Generative Priors and Contact-induced Constraints
}

\author{Minghan Zhu$^{*1,2}$, Zhiyi Wang$^{*2}$, Qihang Sun$^{2}$, Maani Ghaffari$^{2}$, and Michael Posa$^{1}$
\thanks{*These authors contributed equally to this work. }
\thanks{$^{1}$GRASP Laboratory, University of Pennsylvania, Philadelphia, PA 19104
        {\tt\small \{minghz, posa\}@seas.upenn.edu}
        }%
\thanks{$^{2}$University of Michigan, Ann Arbor,
        MI 48109
        {\tt\small \{minghanz, jerrywzy, qhsun, maanigj\}@umich.edu}
        }%
}

\usepackage[style=ieee,citestyle=numeric-comp,maxbibnames=1,minbibnames=1]{biblatex}
\newcommand{\new}[1]{\textcolor{black}{#1}} 
\begin{document}

\maketitle
\thispagestyle{empty}
\pagestyle{empty}

\begin{abstract}

Object geometry is key information for robot manipulation. 
Yet, object reconstruction is a challenging task because camera observations are partial due to occlusions. The scene may not offer the flexibility for a robot to alter its viewpoint to obtain a full observation of the object of interest. In this paper, we leverage two extra sources of information to reduce the ambiguity of vision signals under occlusion. 
First, generative models learn priors of the shapes of commonly seen objects, allowing us to make reasonable guesses of the unseen part of geometry. Second, contact information, which can be obtained from videos and physical interactions, provides sparse constraints on the boundary of the geometry. We combine the two sources of information through contact-guided 3D generation. The guidance formulation is inspired by drag-based generative image editing. \new{We explore different guidance strategies and highlight the importance of short gradient paths for guided generation.} Experiments on synthetic and real-world data show that our approach improves the object reconstruction compared to pure 3D generation and contact-based optimization methods. Project page: \href{https://contactgen3d.github.io/}{https://contactgen3d.github.io/}
\end{abstract}

\section{INTRODUCTION}
While vision is a primary perception modality, it fundamentally provides only partial scene information, making environment state estimation a classic ill-posed problem for robotics. In object manipulation, for instance, recovering essential geometry and pose is often hindered by self-occlusion and occlusions by other objects and the robot itself. Furthermore, changing viewpoints to resolve these occlusions is frequently infeasible in dense environments where workspace and visibility are strictly limited. 
This long-standing challenge has steered some robotic research efforts to end-to-end approaches \cite{chi2023diffusion,brohan2022rt}, where policies and actions are generated directly from raw observations, but they still suffer from limited generalizability and interpretability. 

The problem of object reconstruction from partial observations may be better solved
if we go beyond the raw pixels and leverage other sources of information.
We can obtain knowledge about the geometry of an object from its contacts with the environment. Contacts can only happen on the boundary of the geometry, thus imposing constraints on the object shape. 
\new{Consider, for example, a robot exploring objects in a cluttered kitchen cabinet. Items in the back are heavily occluded from all viewpoints, but as the robot pushes and nudges within the cabinet, contacts provide critical information.}
On the other hand, by learning priors of object geometry from large datasets, generative models can also build a full object model from a partially occluded image. 
Interestingly, the information from contact-induced constraints and data-driven shape priors is usually complementary, as the ``inpainting'' of unobserved geometry by the generative model often has multiple solutions, and the sparse contact constraints reduce the ambiguity. 

\begin{figure}
    \centering
    \includegraphics[width=0.8\linewidth]{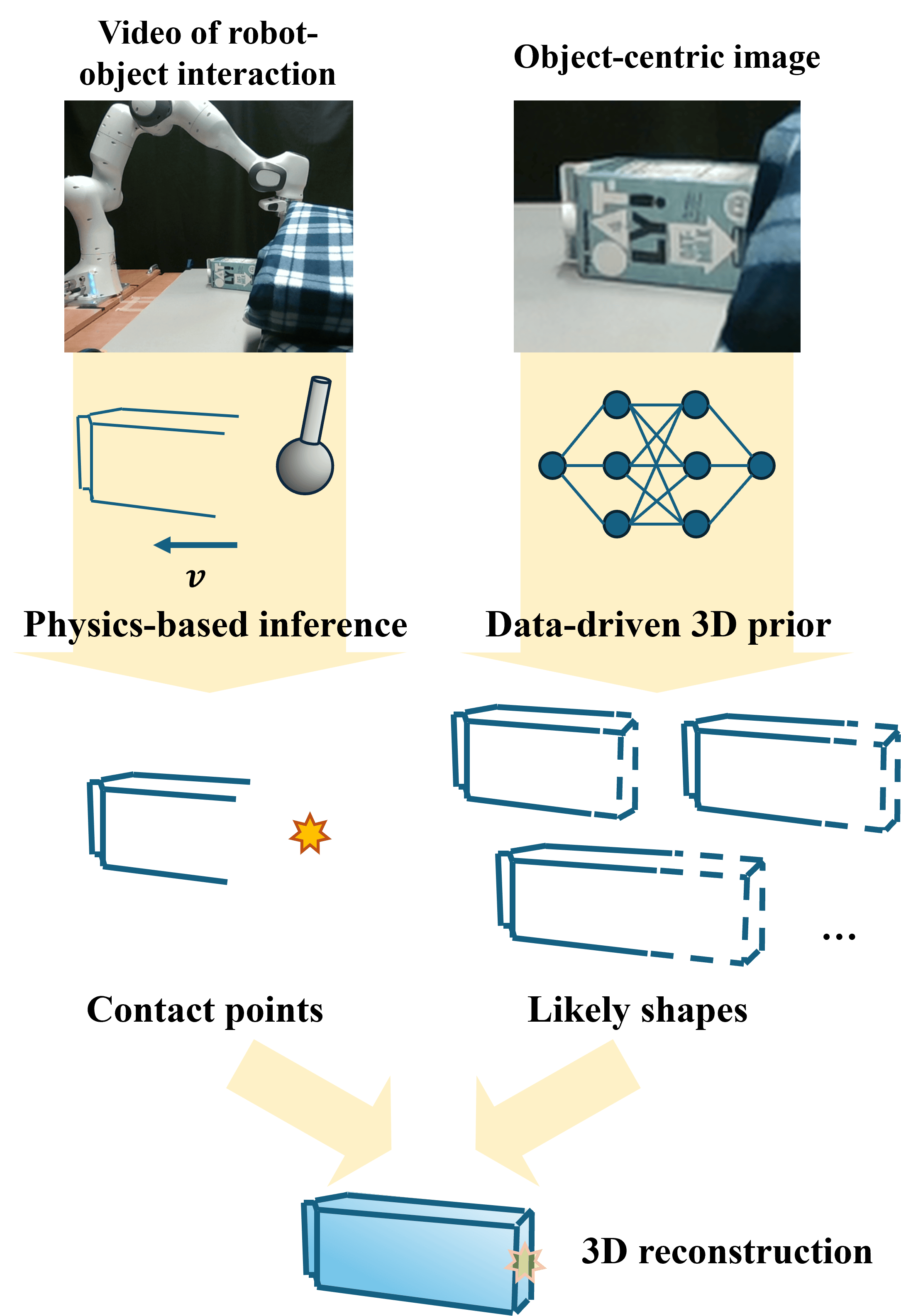}
    \caption{This work develops a novel framework for 3D object reconstruction under occlusion by integrating data-driven 3D priors and physics-based contact information through guided generation. The two perspectives bring complementary insights that lead to high-quality and accurate 3D reconstruction. }
    \vspace{-6pt}
    \label{fig:title}
\end{figure}

In this paper, we tackle the problem of visual object reconstruction under occlusion by combining data-driven 3D priors and contact-induced constraints. A high-level overview is depicted in Fig. \ref{fig:title}. 
We use the paradigm of image-conditioned object generation to harness the powerful generative prior of object shapes. Specifically, we use a flow-matching 3D generative model specifically designed to work with occluded 2D images, Amodal3R \cite{wu_amodal3r_2025}, as our base model for 3D generation. 
\new{On the other hand, we model contact information as a sparse set of contact points. In the real-world experiments, we employ Vysics \cite{bianchini2025vysics} to estimate contact points from very brief (about 10 seconds long) videos and physical interactions. }

We formulate the fusion as 3D generation with training-free contact guidance, where a contact-based energy function is formulated based on the contact points and used to guide the inference of the flow-matching generative model. Specifically, we are inspired by drag-based editing in 2D image generation, and propose a drag-based 3D energy function for contact guidance. 
In this way, we can build the object geometry model from heavily occluded visual observations with high quality and good metric accuracy.  

We summarize the core contributions of this work as follows:
\begin{itemize}
    \item We propose an object reconstruction framework that improves the accuracy of generative 3D priors with sparse physical cues when vision is partially occluded.
    \item We develop a guidance method for native 3D generation using a drag-based energy function. 
    \item Our method achieves superior object reconstruction in synthetic and real-world data, highlighting the importance of synthesizing data-driven and physics-driven insights. 
\end{itemize}

\section{RELATED WORK}
\subsection{3D Object Reconstruction}
There are different paradigms that can be applied to the problem of object reconstruction. Object SLAM \cite{wen2023bundlesdfneural6doftracking, wen2024foundationpose} approaches rely on low-level tracking and shape-pose optimization, achieving good efficiency and generalizability, but lack robustness against visual signal degradation. In this paper, we mainly discuss object reconstruction by 3D generation, in which data-driven priors are used to build the complete geometry (and sometimes appearance) model of an object from one or a few images. 

The development of 3D generative methods is largely inspired and supported by the progress in 2D image generative models. The denoising steps in 2D diffusion \cite{rombach2022highresolutionimagesynthesislatent} can be interleaved with the regression \cite{szymanowicz2023viewsetdiffusion0imageconditioned3d, xu2023dmv3ddenoisingmultiviewdiffusion} or optimization \cite{poole2022dreamfusiontextto3dusing2d} of 3D representations to reconstruct the 3D target. Moreover, multi-view image generation from a single image input 
\cite{liu_zero-1--3_2023,wang2023imagedreamimagepromptmultiviewdiffusion}
provides strong 3D priors. Many works connect multi-view image generation with multi-view 3D reconstruction for 3D generation from a single image
\cite{liu2023one2345singleimage3d, li2023instant3dfasttextto3dsparseview,long2023wonder3dsingleimage3d}, or repurpose multi-view image generation capability for pixel-aligned 3D representation generation \cite{lin_diffsplat_2025}. On the other hand, 3D generative priors can also be learned natively in 3D without relying on image generative priors. Earlier attempts
\cite{mittal_autosdf_2022,cheng_sdfusion_2023}
are limited by the relatively small size of 3D datasets \cite{chang2015shapenetinformationrich3dmodel}. With larger datasets of 3D objects 
\cite{deitke2023objaverse} 
being released, 3D generation models become much more powerful. Some \cite{zhang2024claycontrollablelargescalegenerative,zhang_3dshape2vecset_2023,zhao_hunyuan3d_2025} simply use implicit learnable tokens as the latent features for 3D generation, while others \cite{xiang_structured_2025,chen20253dtopiaxlscalinghighquality3d,lan2025gaussiananythinginteractivepointcloud} use explicit 3D primitives in the latent space for better interpretability and controllability. We used a model based on explicit latent to formulate the contact-induced energy function more easily. 

\subsection{Occlusion-aware 3D Reconstruction}
Despite the significant progress in 3D generation, most such models are trained to take an unoccluded image of the target as input, and their performance degrades significantly when occlusion is present. To overcome this limitation, recent research has explored three main directions. 

\subsubsection{Completing occluded areas in 2D}

In this paradigm, the occluded single-view image is first processed by a 2D amodal segmentation and completion model 
\cite{ozguroglu_pix2gestalt_2024, chen_using_nodate}
or an image inpainting model 
\cite{xie2022smartbrushtextshapeguided, lugmayr_repaint_2022}
to recover the missing regions in the image space. 
The completed image is then provided to a 3D generative model for reconstruction. 
A representative example is SceneComplete \cite{agarwal_scenecomplete_2024}, which integrates BrushNet \cite{ju_brushnet_2024} with InstantMesh \cite{xu_instantmesh_2024}. CAST \cite{yao2025castcomponentaligned3dscene} employs DINOv2's masked-autoencoding capability \cite{oquab2023dinov2} to complete the occluded area in the image latent space. 
However, the effectiveness of this pipeline is constrained by the quality of the 2D amodal completion or inpainting. It is also less straightforward to inject 3D guidance into the 2D generation process, which is important for our work. 

\subsubsection{Shape completion from partial point clouds}  

\new{It is possible to use partial point clouds as the input observation to 3D generative models for shape completion
\cite{mittal_autosdf_2022, cheng_sdfusion_2023, chou_diffusion-sdf_2023,zhang_3dshape2vecset_2023}%
. However, a lot of them require the input points to reside in the canonical object frame, which is usually unavailable in real-world use cases. They also typically require extra image, text, or category condition, as a partial point cloud alone may not carry enough information to recover the whole object. Another paradigm is point cloud completion \cite{yuan2019pcnpointcompletionnetwork, yu2021pointrdiversepointcloud, chen2023anchorformer}, some of which do not require canonical-frame input \cite{humt2023shape, wang2025learning}, but they are typically trained on a smaller dataset \cite{chang2015shapenetinformationrich3dmodel} than the 3D generative models, limiting their performance and generalizability. Therefore, we use images as the input for our 3D reconstruction task. }

\subsubsection{Fine-tuning 3D generative models for occlusion}  

Amodal3R \cite{wu_amodal3r_2025} fine-tunes a native 3D generation framework, TRELLIS \cite{xiang_structured_2025}, with occlusion and visibility masks as additional input modalities, enabling robust handling of occlusions in single-view reconstruction.

\subsection{Adding Controllability to Pretrained Generative Models}
Here, we discuss three strategies for adding controllability to pretrained generative models. 
First, fine-tuning. Technologies like ControlNet \cite{zhang2023adding} and adapters \cite{mou2023t2iadapterlearningadaptersdig} allow fine-tuning a pretrained model to take in new modalities, usually spatial control signals such as edge maps and poses, without destroying the original representations. Second, training-free guidance, which does not require training the generative model with the condition. The guidance signal is formulated as an energy function to represent the compatibility between the generated states and the task-specific needs. In diffusion or flow-matching models, the goal is to find a delta step in each denoising iteration to steer the output to satisfy the task-specific needs. There has been extensive literature on diffusion guidance 
\cite{bansal2023universal,yu2023freedom,chung2022diffusion}
and recently on flow-matching
\cite{feng2025guidance}
as well. Third, we can also treat the controllable generation problem as an editing problem, where uncontrolled generation is first performed, and then we apply generative editing 
\cite{mou2023dragondiffusion,shi2024dragdiffusionharnessingdiffusionmodels}
to make the outcome follow our specifications. In this paper, we incorporate the idea of editing in building a training-free guidance for contact-conditioned 3D generation. 

\subsection{Physics-based Reasoning in 3D Reconstruction}
As discussed in the introduction, physics-based reasoning helps reduce the ambiguity in visual signals and build a physically plausible reconstruction. Many efforts assume the objects to be stable and static and use optimization to prevent penetration and unstable falling \cite{agnew2021amodal,ni2024phyrecon,yao2025cast}. In \cite{bianchini2025vysics,yang2025twintrack,song2018inferring}, contact is analyzed in dynamic scenes to improve 3D reconstruction. Estimated physics properties may also be used to generate physically realistic videos
\cite{chen2025physgen3d,jiang2025phystwin}. 

\section{PRELIMINARIES}
\subsection{3D Generation: TRELLIS and Amodal3R}
Amodal3R \cite{wu_amodal3r_2025} is a 3D generation model taking occluded images and masks as input.  It is fine-tuned on TRELLIS \cite{xiang_structured_2025}, a flow-matching model capable of image-to-3D and text-to-3D generation. Our framework uses Amodal3R as the base model to provide 3D priors. Thus we will first introduce TRELLIS and Amodal3R briefly. 

TRELLIS contains two stages for 3D generation: sparse structure generation and structured latent generation, which are in charge of the generation of the overall geometry and the local details and appearance, respectively. The sparse structure is an occupancy voxel grid of resolution~64, $s\in \mathbb{R}^{64^3}$, representing the geometry in a unit cube. The structured latent is feature vectors attached to each occupied voxel, which can be decoded into 3D Gaussians or FlexiCube \cite{Shen_2023} representations of resolution $256^3$ for mesh generation. The first stage output resolution is high enough for meaningful contact-induced guidance, thus we mainly focus on the first-stage guided generation in this paper, and keep the second stage as is. In the first stage, the latent space for sparse structure generation is a dense 3D voxel feature map with downsampled spatial resolution, $x\in \mathbb{R}^{16^3\times8}$ (where 16 is the spatial resolution and 8 is the feature dimension), which can be mapped from and to the space of $s$ through a UNet voxel encoder $\mathcal{E}$ and decoder $\mathcal{D}$, respectively. The generation is through flow matching with Diffusion Transformers $\mathcal{T}$ (DiT) \cite{peebles2023scalable}. Amodal3R incorporates a visibility mask, which highlights the visible part of the object of interest, and an occlusion mask highlighting the occluder, as extra input. Please refer to the original papers for more details. 

\subsection{Vysics: Contact Estimation from Video}
In our real-world experiments, contact information comes from Vysics \cite{bianchini2025vysics}, an optimization-based object tracking and 3D reconstruction framework that also infers contact points and associated contact force from a video of an object interacting with a robot and the environment, and uses the estimated contacts to assist in building the object model. 
\new{The set of contact points serves as a sparse representation of regions of high contact force on the surface of the object during the interactions, which could include sliding point-to-surface contacts and surface-to-surface contacts (e.g., a table surface supporting the object). In other words, one should not interpret the number of contact points as the number of required isolated pokes on the object. The contact points can be obtained through more natural and brief interactions. }
In this paper, we use the estimated contact points from Vysics as the guiding signal for the 3D generation under occlusion. 

\subsection{Drag-based Generative Editing}
The operation ``dragging'' originates from 2D image editing, where the goal is to move a point in the image to a different coordinate, while keeping the overall image semantically consistent and spatially reasonable. After the pioneering work DragGAN \cite{pan2023drag}, the drag-based editing is also realized in diffusion models \cite{mou2023dragondiffusion,chen2024mvdrag3d}. On a high level, the drag-based editing in diffusion models can be done by adding noise to an image (as the reference) and denoising. In the denoising process, a drag-based loss encourages the features around the target point to resemble the reference features around the source point, and a content-preservation loss encourages features everywhere else to be consistent with the reference. We will build our guidance in a similar drag-based formulation. 

\new{\section{METHOD}}

\subsection{Problem Formulation}
We formulate the fusion of data-driven 3D priors and sparse contact point data for object reconstruction as a guided flow-matching problem. Flow-matching models define a vector field $v_t$ that establishes a probability path $p_t(x_t)$ from an initial base distribution $p_1$ (noise) to a target distribution $p_0$ (3D objects). By drawing a sample $x_1 \sim p_1(x_1)$ and integrating the ordinary differential equation (ODE) $dx_t = v_t(x_t)dt$, we generate samples $x_0$ that conform to the target distribution $p_0(x_0)$. The flow-matching model learns the vector field $v_t(x_t)=\mathcal{T}(x_t;t)$.

We extend this to a guided setting where the goal is to steer the probability path toward a conditional distribution $p_t(x_t|c)$, where $c$ represents the condition, i.e., the set of specified contact points in this problem. This is achieved by determining a guidance vector field $g_t(x_t)$ such that the updated vector field, $v'_t(x_t) = v_t(x_t) + g_t(x_t)$, directs the integration from $p_1$ toward a state compatible with the constraints. Following the principle of training-free energy-based guidance \cite{yu2023freedom}, we define an energy function $J(x_t; c)$ that represents the local discrepancy between the current state and the contact constraints. The value of $J$ should be small when $x_t$ is more compatible with the contact condition $c$, and large otherwise. The guidance vector is then computed as the gradient of this energy with respect to the current state:
\begin{equation}
g_t(x_t) = -\lambda_t \nabla_{x_t} J(x_t; c),
\end{equation}
where $\lambda_t$ is the step length, typically a hyperparameter. 
$J$ can be instantiated in various ways. Considering that $x_t$ is a noisy latent state with no clear geometric meaning, it may be tempting to map $x_t$ to the noise-free occupancy $s_0$ to measure the compatibility with given contact points. However, the drag-based guidance formulation, which converts the point condition into a feature-similarity loss, offers more flexibility in designing the energy function $J$, as detailed in Sec. \ref{sec:dragging}. 

\subsection{Drag-based Contact Guidance}\label{sec:dragging}
The objective of contact guidance is to steer the generative process such that the resulting 3D geometry satisfies sparse physical constraints while remaining on the manifold of plausible shapes. We define the guidance energy $J$ as the local discrepancy between the generated representation around a contact point $p^c\in c$ and a reference representation around some source location to ``drag'' from. To obtain this source, we first perform an unguided pass of the flow-matching model to generate a reference shape $s^*_0$, which should be a plausible shape given the image but unaware of the contact condition. Then we identify the nearest surface point $p^s$ on the reference shape as the source point for each target contact point $p^c$.

Given the source and target point pairs and the goal of matching the two neighborhoods, we now need to decide what feature space to measure the similarity and instantiate the guidance. We explore three options, representing different trade-offs between spatial resolution, numerical stability, and computational complexity. Fig. \ref{fig:guide_options} illustrates the three guidance options. 

\subsubsection{Voxel-Space Guidance}
\begin{figure}
    \centering
    \includegraphics[width=\linewidth]{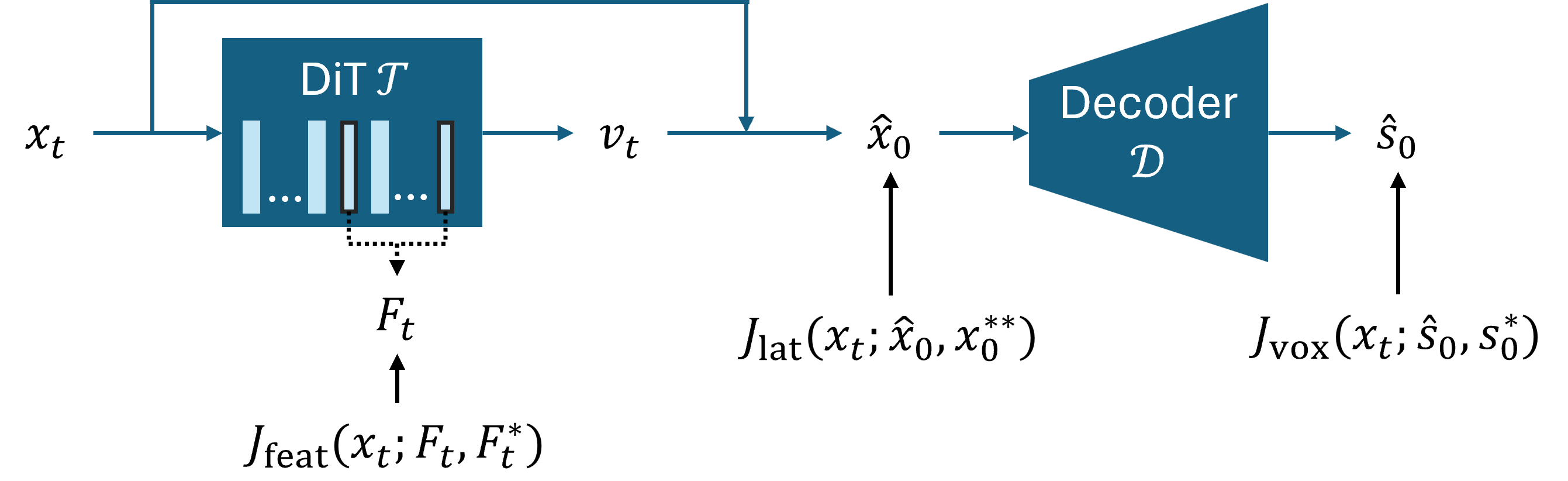}
    \caption{The three guidance options visualized in one denoising step. }
    \vspace{-6pt}
    \label{fig:guide_options}
\end{figure}

The most direct approach is to measure compatibility in the decoded occupancy space. The loss is computed as:
\begin{equation}
J_{\text{vox}} = \frac{1}{|\mathcal{N}_\epsilon|} \sum_{\Delta p \in \mathcal{N}_\epsilon} | \hat{s}_0(p^c + \Delta p) - s_0^*(p^s + \Delta p) |^2,
\end{equation}
where $\mathcal{N}_\epsilon$ is a 3D neighborhood, and $\hat{s}_0 = \mathcal{D}(\hat{x}_0) = \mathcal{D}(x_t - v_t t)$ is the decoded noise-free occupancy estimated from the current noisy latent state $x_t$.\footnote{Here we assume affine conditional probability path for the flow model, which is true for TRELLIS. } While this offers high spatial resolution, the gradient path of backpropagation ($\hat{s}_0 \rightarrow \mathcal{D} \rightarrow \hat{x}_0 \rightarrow v_t \rightarrow \text{DiT} \rightarrow x_t $, see Fig.~\ref{fig:guide_options}) is deep and prone to accumulating gradient noise.

\subsubsection{Latent-Space Guidance with Sub-coarse-grid Shifting}
To bypass the backpropagation through the decoder, one can guide the denoised latents $\hat{x}_0$ directly. However, the latent grid is coarser ($16^3$), meaning a standard drag would only be precise to integer coarse-grid units. To achieve ``sub-coarse-grid'' precision, we shift the reference occupancy $s^*_0$ by integer units in the $64^3$ fine grid such that the offset between the source and target points is of integer units in the coarse latent space. This shifted occupancy is re-encoded to a target latent $x^{**}_0$, and we apply a cosine-based drag loss:
\begin{equation}
J_{\text{lat}} = \frac{1}{|\mathcal{N}_\epsilon|} \sum_{\Delta p \in \mathcal{N}_\epsilon} D\left(\hat{x}_0(p^c + \Delta p), x^{**}_0(p^s + \Delta p) \right),
\end{equation}
where $D(\cdot)$ is the inverse normalized cosine similarity:
$$D(x,y)=\frac{1}{0.5 \cdot \frac{x\cdot y}{\|x\|\|y\|} + 0.5}.$$
While this provides contact-point following of the same spatial resolution as the voxel-space guidance, the backpropagation from $\hat{x}_0$ to $x_t$ still introduces artifacts, and the re-encoding needs extra computation. 

\subsubsection{Feature-Space Guidance}
Our recommended approach applies guidance directly to the intermediate feature layers $F_t$ of the DiT $\mathcal{T}$ that takes $x_t$ as input. We target the middle and final blocks of the DiT to capture a balance of semantic and geometric structure. This is similar to the practice in DragonDiffusion \cite{mou2023dragondiffusion}, which applies the drag-based image editing loss on the intermediate layers of the denoising UNet. Our guidance energy is defined as:
\begin{equation}
J_{\text{feat}} = \frac{1}{|\mathcal{N}_\epsilon|} \sum_{\Delta p \in \mathcal{N}_\epsilon} D\left(F_t(p^c + \Delta p), F^*_t(p^s + \Delta p) \right).
\end{equation}
By guiding the features internally, we shorten the gradient path and avoid the instabilities associated with noise-free state prediction. Although limited to the $16^3$ resolution, this manifold provides the highest stability and best overall 3D reconstruction accuracy, as shown later in Sec. \ref{sec:syn}.

\begin{figure*}
    \centering
    \includegraphics[width=\textwidth,page=2]{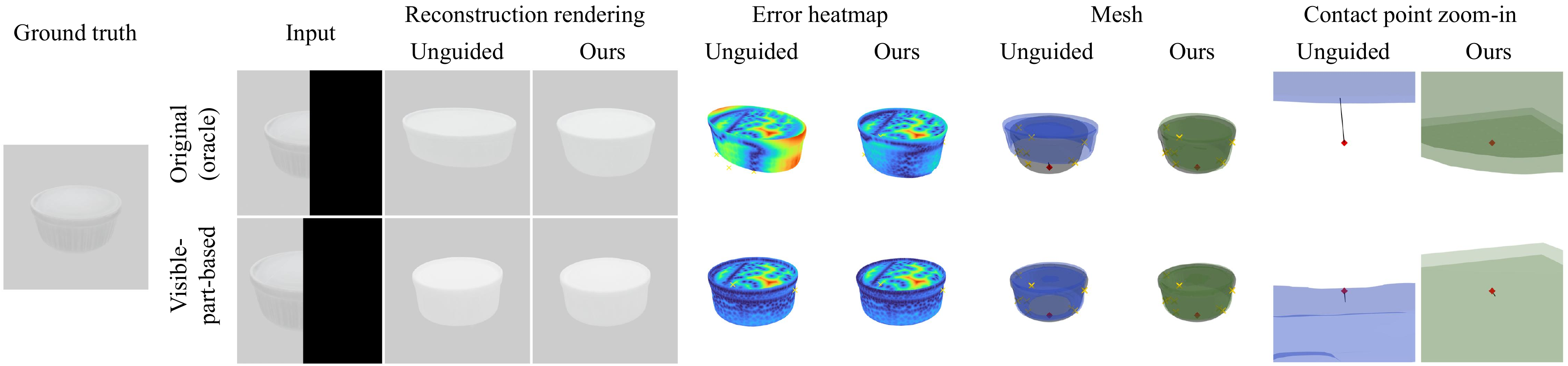}
    \includegraphics[width=\textwidth,page=2]{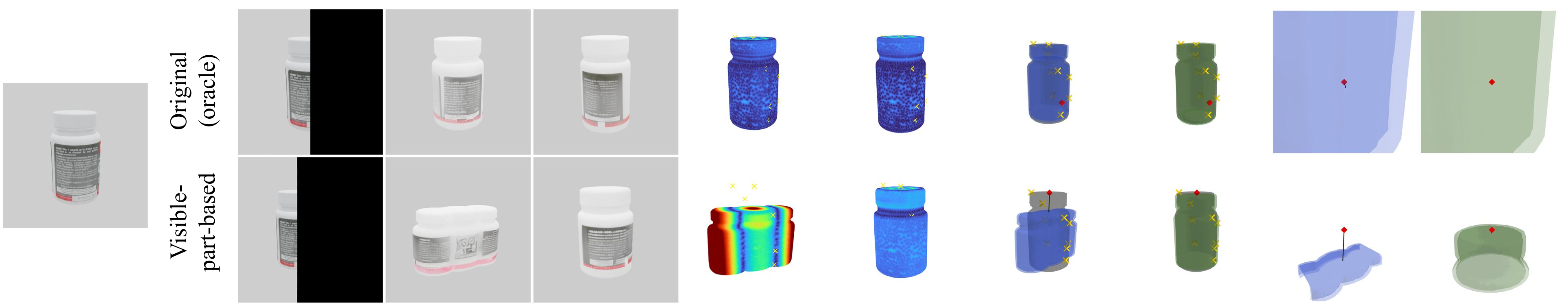}
    \includegraphics[width=\textwidth,page=3]{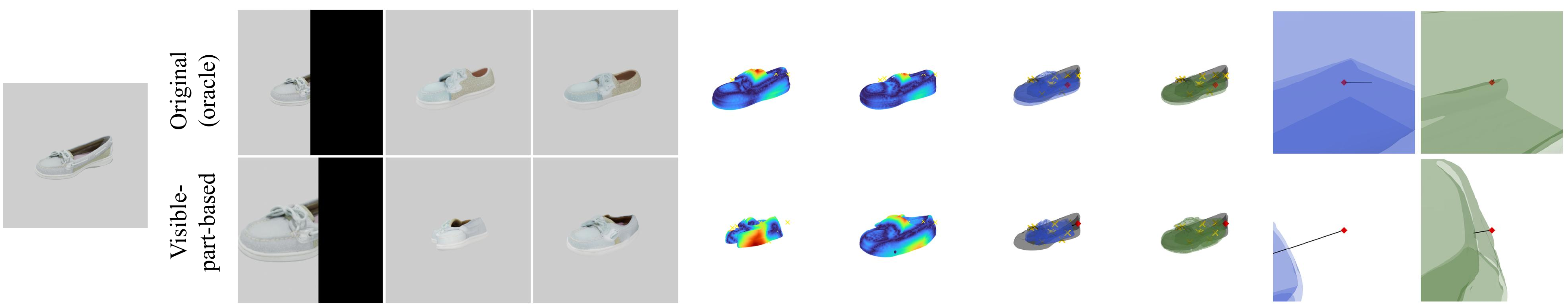}
    \includegraphics[width=\textwidth,page=4]{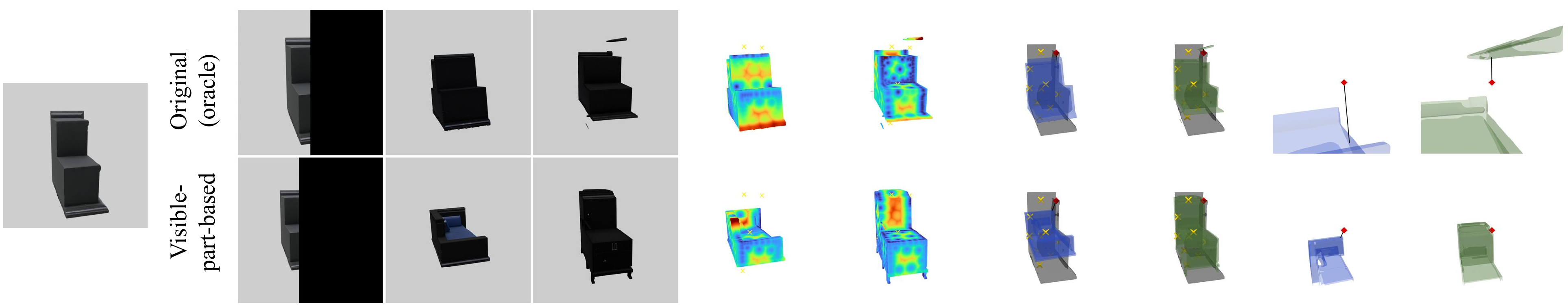}
    \caption{Qualitative comparison of the geometry reconstruction. For each object, we show the reconstruction under two input cropping conditions (see Sec. \ref{sec:exp_syn}). Occlusion masks are shown in black. The error heatmaps depict the one-sided point-wise Chamfer distance from the predicted shape to the ground truth. The mesh view shows predictions in blue (unguided) and green (guided) overlaid on the ground-truth mesh in gray. The contact points are shown in yellow crossings in both error heatmaps and mesh views, with one point in red zoomed-in in the right-most two columns with a black line connecting the contact point to the nearest point on the predicted surface. }
    \label{fig:heatmap}
\end{figure*}

\begin{table*}[]
    \centering
    \resizebox{\textwidth}{!}{

\begin{tabular}{llcccccccccccc}
\toprule
 \multirow{2}{*}{\makecell[l]{Input \\ condition}}&\multirow{2}{*}{Guidance method}               & \multicolumn{2}{c}{Chamfer Distance ↓}           & \multicolumn{2}{c}{F@0.01 ↑}    & \multicolumn{2}{c}{F@0.02 ↑}    & \multicolumn{2}{c}{F@0.05 ↑}    & \multicolumn{2}{c}{Contact Distance ↓}    & \multicolumn{2}{c}{RGB-LPIPS ↓}   \\
                       && Mean            & Median           & Mean           & Median         & Mean           & Median         & Mean           & Median         & Mean            & Median          & Mean           & Median         \\ \midrule
 \multirow{4}{*}{\makecell[l]{Original \\ (oracle)}} & Unguided \cite{wu_amodal3r_2025}        & 0.0390& 0.0329
& 0.380& 0.350& 0.658& 0.677& 0.910& 0.968
& 0.0169& 0.0108
& \textbf{0.154} & \textbf{0.135}
\\
 &Voxel ($J_\text{vox}$)       & 0.0414& 0.0380
& 0.360& 0.323& 0.629& 0.624& 0.901& 0.955
& 0.0165& 0.0112
& 0.303& 0.302
\\
 &Latent ($J_\text{lat}$)         & 0.0411& 0.0381
& 0.354& 0.306& 0.626& 0.599& 0.904& 0.948
& \textbf{0.0149}& \textbf{0.0100}
& 0.175& 0.163
\\
 & Feature  ($J_\text{feat}$, \textit{ours})    & \textbf{0.0375}& \textbf{0.0309}
& \textbf{0.384} & \textbf{0.379} & \textbf{0.665} & \textbf{0.693} & \textbf{0.918} & \textbf{0.976}
& \textbf{0.0149}& 0.0114
& 0.156& 0.137
\\ \midrule
  \multirow{4}{*}{\makecell[l]{Visible- \\ part- \\ based}} & Unguided \cite{wu_amodal3r_2025}        
& 0.0475& 0.0429
& 0.330& 0.288& 0.585& 0.563& 0.868& 0.923
& 0.0213& 0.0152
& \textbf{0.161}&0.145
\\
  &Voxel ($J_\text{vox}$)       
& 0.0473& 0.0415
& 0.327& 0.294& 0.583& 0.564& 0.868& 0.935
& 0.0207& 0.0139
& 0.303&0.302
\\
  &Latent ($J_\text{lat}$)         
& 0.0467& 0.0420
& 0.326& 0.280& 0.584& 0.563& 0.871& 0.916
& 0.0185& 0.0135
& 0.175&0.163
\\
  & Feature ($J_\text{feat}$, \textit{ours})        & \textbf{0.0430}& \textbf{0.0374}& \textbf{0.345}& \textbf{0.302}& \textbf{0.613}& \textbf{0.608}& \textbf{0.892}& \textbf{0.960}& \textbf{0.0179}& \textbf{0.0127}& 0.162& \textbf{0.139}\\ \bottomrule
\end{tabular}
    }
    \caption{Quantitative evaluation of 3D reconstruction under occlusion on the GSO \cite{downs2022googlescannedobjectshighquality} dataset, with two input conditions. ↓ means smaller is better, ↑ means larger is better. The best is highlighted in bold font. The objects are normalized in a unit cube. }
    \label{tab:gso}
\end{table*}

\section{EXPERIMENTS}
\subsection{Shared setup for both experiments}
\new{The goal of our experiments is to demonstrate that sparse physical cues can significantly refine a generative 3D prior when visual observations are heavily occluded. We evaluate this on a large-scale synthetic dataset and a real-world robotic interaction dataset. In both settings, the model receives a single 2D image of the target object with partial occlusion as visual input for the 3D generation. 
In the synthetic experiment, the points are sampled from the ground-truth object surface. In the real-world experiment, the points are estimated by Vysics \cite{bianchini2025vysics}. }

We use the same set of hyperparameters for the experiments on synthetic and real data.
The flow-matching model takes 12 timesteps in total for inference. 
\new{The 3D neighborhood $\mathcal{N}_{\epsilon}$ for drag-based contact guidance is set to a $21^3$ cube in the fine voxel grid and $5^3$ cube in the coarse feature grid, which roughly correspond to the same neighborhood area. }

\subsection{Implementation details on guidance}
\new{We empirically found that the voxel-space and latent-space guidance is more numerically sensitive and requires careful scheduling of the guidance step size $\lambda_t$. We split the 12 denoising steps into early, middle, and late stages, each with 4 timesteps. $\lambda_t$ is set to 0.2, 1.0, and 0.5, for the early, middle, and late stages, respectively. We run the guidance recurrently \cite{bansal2023universal} three times at each timestep. For the feature-space guidance, we simply use $\lambda_t=1$ and do not apply recurrent guidance. }

\subsection{Experiment on synthetic data}\label{sec:exp_syn}
The synthetic experiment utilizes the Google Scanned Objects (GSO) \cite{downs2022googlescannedobjectshighquality} dataset, 
a collection of 3D-scanned household items.
We randomly select 200 objects and render them with artificial occlusions by masking out half of the image. We generate synthetic contact points by randomly sampling 10 points from the unobserved parts of the object surface. Since TRELLIS \cite{xiang_structured_2025} is trained on object-centered renderings, under heavy occlusion Amodal3R \cite{wu_amodal3r_2025} tends to assume the full shape is centered too. Our masked GSO renderings are themselves object-centered, which tends to help the unguided baseline guess the occluded half more accurately from this assumption alone, an advantage unavailable in practice, where the full extent is unknown beforehand. To control for this, we evaluate two input conditions on GSO: an \textit{Original (oracle)} condition using the ground-truth object placement and scale, and a \textit{Visible-part-based} condition, where the object is re-scaled and re-centered using only the visible half heuristically, mirroring what is achievable under real-world occlusion. 
Concretely, the crop scales the visible mask's maximal side length to a fixed ratio of the image, and centers it with a left offset, shifted as needed to keep the whole mask inside the frame. This heuristic is not intended to be a research contribution, but a minimal engineering effort to crop reasonably without knowledge of the full shape.

\begin{figure}
    \centering
    \includegraphics[width=\linewidth]{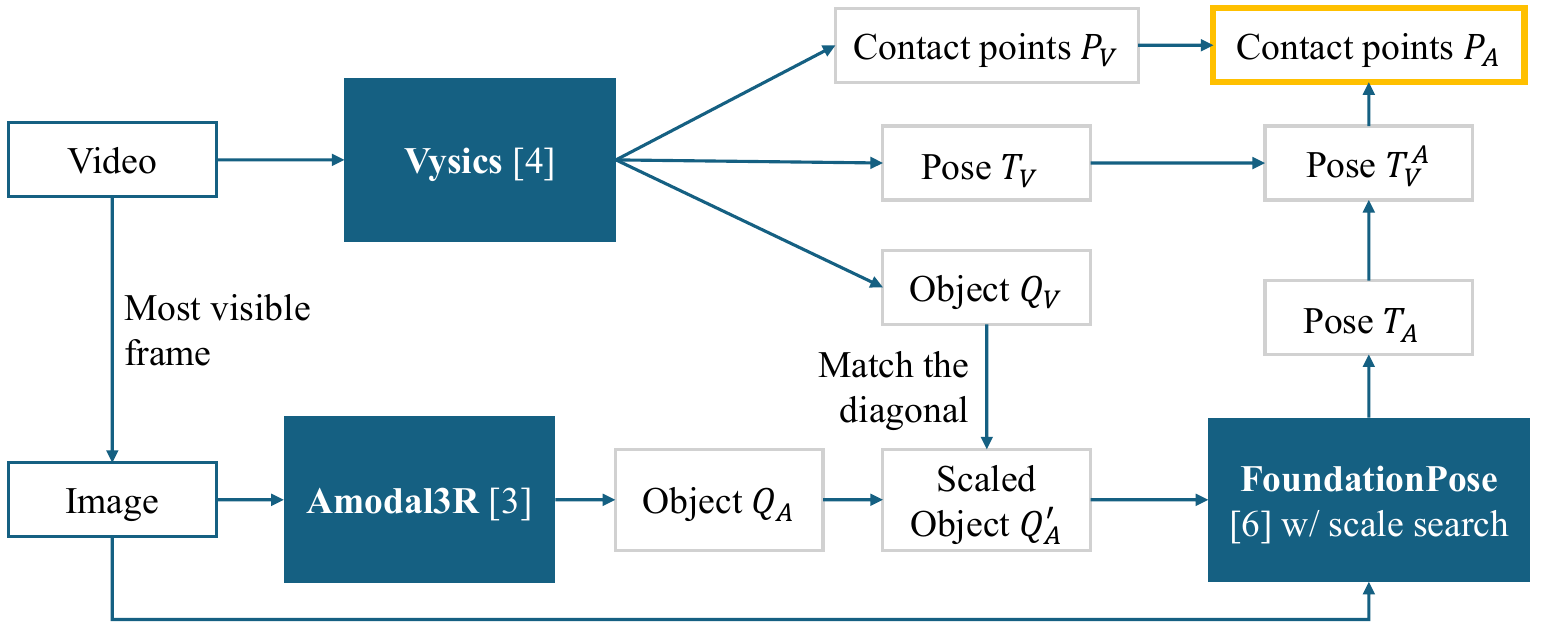}
    \caption{The process of converting contact points estimated by Vysics to the prediction space of Amodal3R for contact-guidance in real-world experiments. $T_V, T_A$ are the poses of objects $Q_V, Q_A$ in the camera frame. $T^A_V$ converts between the two object frames. }
    \vspace{-6pt}
    \label{fig:system}
\end{figure}

\subsection{Experiment on real-world data}
Real-world validation uses the Vysics \cite{bianchini2025vysics} dataset featuring a Franka robot interacting with tabletop objects through pushing, pivoting, and toppling. The data includes RGBD videos with partially occluded objects, the corresponding masks, the robot trajectories, and the ground-truth object mesh for evaluation. There are multiple videos of interactions, each approximately 10 seconds long, for every object, and the inference is on each video independently. Contact points are derived from each video of interaction, while the generative model uses a single image frame from the video when the area of the visible object mask is the largest. We also crop the image around the visible mask using the same heuristic as in the synthetic experiment (Sec. \ref{sec:exp_syn}) to make the visible part of the object roughly centered and occupy the major part of the image. We do not have privileged information about the hidden part of the object when cropping the images.

In order to convert the contact points to Amodal3R's prediction frame, we first run the unguided Amodal3R to obtain an initial reconstructed mesh. The mesh scale is initialized by matching the diagonal length of the best-fitting oriented bounding box between Amodal3R's predicted mesh and Vysics' reconstructed mesh using Open3D \cite{zhou2018open3d}. Then we align the mesh to the RGBD image using FoundationPose \cite{wen2024foundationpose} with scale refinement. Vysics also yields the object's tracked pose in each RGBD frame. Therefore, the contact points in Vysics' object frame can be converted to Amodal3R's object frame. We then run a second run of Amodal3R generation with the guidance of contact points. 
Vysics typically yields hundreds of contact points that could be redundant for explaining the contact dynamics. We subsample 20 points using farthest point sampling for Amodal3R's guidance. An overview of this process is depicted in Fig. \ref{fig:system}. 

\subsection{Results on synthetic data}\label{sec:syn}
Table \ref{tab:gso} shows the quantitative results of 3D reconstruction under occlusion on the GSO dataset \cite{downs2022googlescannedobjectshighquality}. In terms of the 3D evaluation, we measure the Chamfer distance and the F-score of the points on the predicted shape at different distance thresholds. We also report contact distance as the distance from the specified contact points to the reconstructed mesh to show the adherence of the generation results to the contact condition. Furthermore, we report the LPIPS score \cite{zhang2018unreasonable} of the rendered RGB images, in order to evaluate the quality of the generated objects. 
\new{Comparing the two input conditions, all methods achieve better 3D accuracy under the \textit{Original (oracle)} condition than under the \textit{Visible-part-based} condition, confirming that the position/scale prior leaks shape information when the object is rendered at its ground-truth placement. Among the three guidance strategies, feature-space guidance is consistently the most accurate, and it is the only strategy that outperforms the unguided baseline in both input conditions. Under the oracle condition, voxel-space and latent-space guidance strategies underperform the unguided baseline while feature guidance slightly outperforms, since the baseline already benefits from the centering shortcut, leaving guidance less room to help. In the \textit{Visible-part-based} condition, however, all three guidance strategies show a consistent improvement in Chamfer distance over the unguided baseline, showing that guidance refines the shape using the contact evidence itself, rather than relying on the centering prior. On the other hand, the unguided baseline still tends to achieve better or comparable LPIPS scores than the guided results across both conditions, suggesting that our guidance can introduce visual artifacts, a limitation also visible for the last object in Fig.~\ref{fig:heatmap}. }

Fig. \ref{fig:heatmap} shows qualitative examples of the 3D reconstruction on the GSO dataset under both input conditions, comparing the unguided and feature-guided predictions. The error heatmaps show that the guided reconstruction has a smaller error to the ground-truth shape. The zoomed-in mesh views further show that the predicted surface moves closer to the specified contact points after guidance, illustrating the local geometric improvement induced by the contact guidance.

\begin{table}[]
    \centering
    \setlength{\tabcolsep}{2pt}
    \resizebox{\columnwidth}{!}{
    \begin{tabular}{lcccccccc}
    \toprule
Categories   & bakingbox     & bottle        & egg           & oatly         & milk          & styrofoam     & toblerone     & all           \\ \midrule
BundleSDF \cite{wen2023bundlesdfneural6doftracking} & 3.84          & 2.65          & 3.70          & 2.45          & 3.17          & 2.55          & 2.44          & 2.98          \\
Vysics \cite{bianchini2025vysics}    & 1.83          & 1.36          & 1.05          & 1.25          & 1.53 & 1.45 & 1.02 & 1.45          \\
Amodal3R \cite{wu_amodal3r_2025}  & 1.12          & \textbf{1.32} & 0.85          & 1.13          & \textbf{1.42}          & 1.26          & 0.89          & 1.18          \\
Ours      & \textbf{1.07} & 1.54          & \textbf{0.82} & \textbf{0.77} & 1.60          & \textbf{1.13}          & \textbf{0.83}          & \textbf{1.11} \\ \bottomrule
\end{tabular}
    }
    \caption{Chamfer distance of the meshes reconstructed under occlusion on the Vysics \cite{bianchini2025vysics} real-world dataset. Unit: cm. }
    \vspace{-6pt}
    \label{tab:vysics}
\end{table}

\begin{figure}
    \centering
    \includegraphics[width=\linewidth]{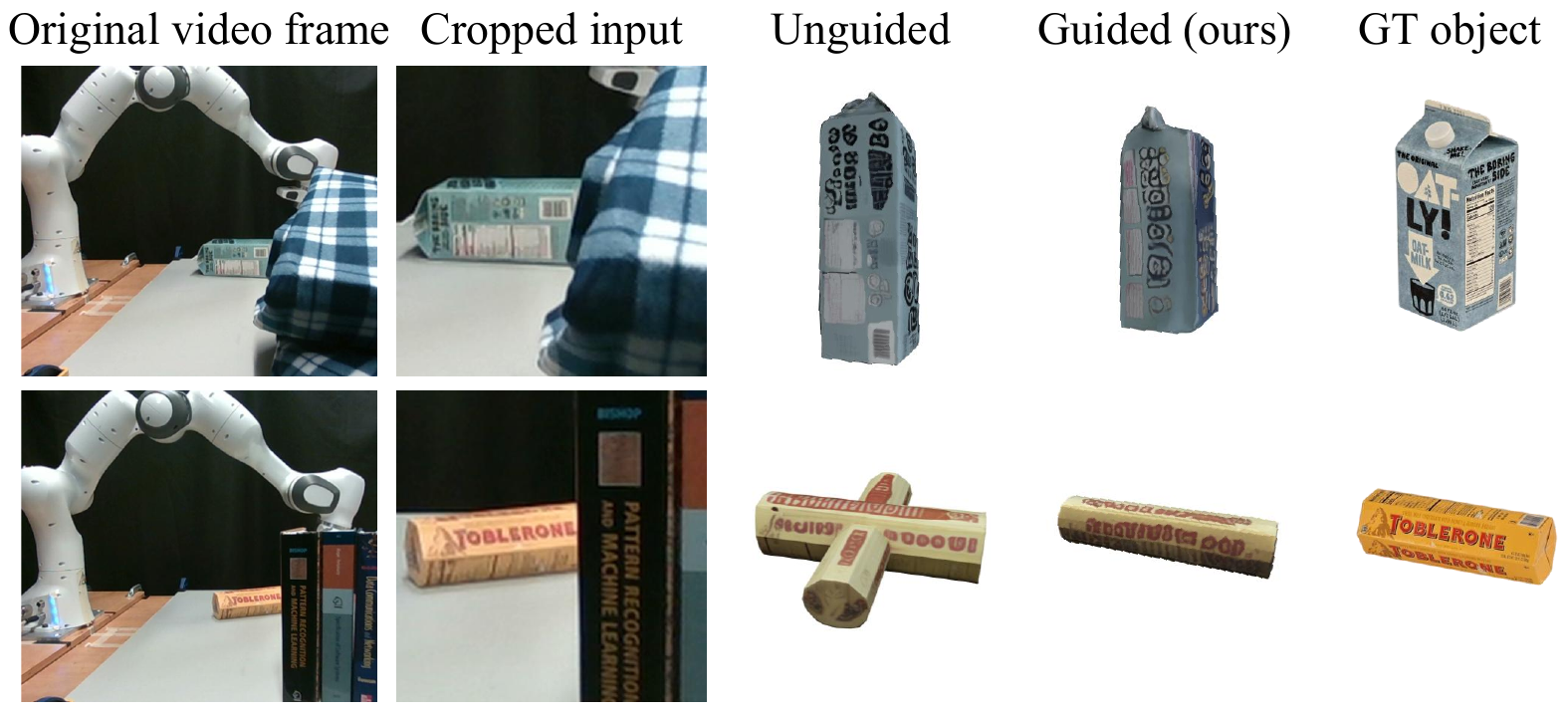}
    \caption{Qualitative examples of 3D reconstruction on Vysics data. The top row is a successful case, and the bottom row is a failure case. }
    \vspace{-6pt}
    \label{fig:vysics}
\end{figure}

\subsection{Results on real-world data}
Table \ref{tab:vysics} shows the quantitative results of the 3D reconstruction under occlusion in the real-world data provided by Vysics \cite{bianchini2025vysics}. The comparisons include BundleSDF \cite{wen2023bundlesdfneural6doftracking}, representing pure vision-based reconstruction; Vysics \cite{bianchini2025vysics}, representing joint optimization of vision and physics-based reasoning; Amodal3R \cite{wu_amodal3r_2025}, representing visual reconstruction with generative prior knowledge; and ours, which combines vision, generative priors, and physics-based insights of the contacts. We use feature-space guidance in this experiment. Our method achieves the overall best geometric accuracy in terms of Chamfer distance. See Fig. \ref{fig:vysics} for example cases. The guidance helps improve the aspect ratio of the reconstructed objects when a vision model alone cannot infer correctly under heavy occlusion. The guidance also alleviates the multi-head Janus problem \cite{shi2024mvdreammultiviewdiffusion3d,poole2022dreamfusiontextto3dusing2d}, likely due to the asymmetry introduced by the contact points.

\section{CONCLUSIONS}
We explored the integration of data-driven 3D shape priors and physics-driven contact information to tackle the challenging task of 3D object reconstruction under heavy occlusion. 
This approach highlights contact as a novel and exploitable modality that complements vision, harnessing the physical constraints of the environment to enable robots to build accurate object models for manipulation in cluttered real-world settings.
Under the paradigm of guided flow-matching generation, we propose a drag-based contact loss function to guide the dense geometry with sparse contact points. Experimental results on synthetic data and real-world data both show that our framework improves the reconstruction accuracy by incorporating both sources of knowledge. We hope this work inspires more explorations of joining the merits of big data and physics-based insights for building more powerful robot vision systems. 

Our work still has limitations. The guided generation gets close to, but may not exactly land on, the specified contact points, and can introduce artifacts. The drag-based guidance loss assumes that the unguided results have a roughly correct structure, which may not always be true in practice. We will explore solutions to these problems in future work. 




\section*{ACKNOWLEDGMENT}
This work was supported by an NSF CAREER Award under Grant No. FRR-2238480 and the RAI Institute. \mbox{M.~Ghaffari} was supported by AFOSR MURI FA9550-23-1-0400 and AFOSR YIP FA9550-25-1-0224.


\printbibliography

@inproceedings{humt2023shape,
  title={Shape completion with prediction of uncertain regions},
  author={Humt, Matthias and Winkelbauer, Dominik and Hillenbrand, Ulrich},
  booktitle={2023 IEEE/RSJ International Conference on Intelligent Robots and Systems (IROS)},
  pages={1215--1221},
  year={2023},
  organization={IEEE}
}

@inproceedings{wang2025learning,
  title={Learning Generalizable Shape Completion with SIM (3) Equivariance},
  author={Wang, Yuqing and Chen, Zhaiyu and Zhu, Xiao Xiang},
  booktitle={Advances in Neural Information Processing Systems (NeurIPS)},
  year={2025}
}

@inproceedings{peebles2023scalable,
  title={Scalable diffusion models with transformers},
  author={Peebles, William and Xie, Saining},
  booktitle={Proceedings of the IEEE/CVF international conference on computer vision},
  pages={4195--4205},
  year={2023}
}

@article{chi2023diffusion,
  title={Diffusion policy: Visuomotor policy learning via action diffusion},
  author={Chi, Cheng and Xu, Zhenjia and Feng, Siyuan and Cousineau, Eric and Du, Yilun and Burchfiel, Benjamin and Tedrake, Russ and Song, Shuran},
  journal={The International Journal of Robotics Research},
  pages={02783649241273668},
  year={2023},
  publisher={SAGE Publications Sage UK: London, England}
}

@article{brohan2022rt,
  title={Rt-1: Robotics transformer for real-world control at scale},
  author={Brohan, Anthony and Brown, Noah and Carbajal, Justice and Chebotar, Yevgen and Dabis, Joseph and Finn, Chelsea and Gopalakrishnan, Keerthana and Hausman, Karol and Herzog, Alex and Hsu, Jasmine and others},
  journal={arXiv preprint arXiv:2212.06817},
  year={2022}
}

@inproceedings{zhang2018unreasonable,
  title={The unreasonable effectiveness of deep features as a perceptual metric},
  author={Zhang, Richard and Isola, Phillip and Efros, Alexei A and Shechtman, Eli and Wang, Oliver},
  booktitle={Proceedings of the IEEE conference on computer vision and pattern recognition},
  pages={586--595},
  year={2018}
}

@article{zhou2018open3d,
  title={Open3D: A modern library for 3D data processing},
  author={Zhou, Qian-Yi and Park, Jaesik and Koltun, Vladlen},
  journal={arXiv preprint arXiv:1801.09847},
  year={2018}
}

@article{song2018inferring,
  title={Inferring 3d shapes of unknown rigid objects in clutter through inverse physics reasoning},
  author={Song, Changkyu and Boularias, Abdeslam},
  journal={IEEE robotics and automation letters},
  volume={4},
  number={2},
  pages={201--208},
  year={2018},
  publisher={IEEE}
}

@article{yang2025twintrack,
  title={TwinTrack: Bridging Vision and Contact Physics for Real-Time Tracking of Unknown Dynamic Objects},
  author={Yang, Wen and Xie, Zhixian and Zhang, Xuechao and Amor, Heni Ben and Lin, Shan and Jin, Wanxin},
  journal={arXiv preprint arXiv:2505.22882},
  year={2025}
}

@inproceedings{chen2025physgen3d,
  title={Physgen3d: Crafting a miniature interactive world from a single image},
  author={Chen, Boyuan and Jiang, Hanxiao and Liu, Shaowei and Gupta, Saurabh and Li, Yunzhu and Zhao, Hao and Wang, Shenlong},
  booktitle={Proceedings of the Computer Vision and Pattern Recognition Conference},
  pages={6178--6189},
  year={2025}
}

@article{jiang2025phystwin,
  title={Phystwin: Physics-informed reconstruction and simulation of deformable objects from videos},
  author={Jiang, Hanxiao and Hsu, Hao-Yu and Zhang, Kaifeng and Yu, Hsin-Ni and Wang, Shenlong and Li, Yunzhu},
  journal={IEEE International Conference on Computer Vision (ICCV)},
  year={2025}
}

@article{yao2025cast,
  title={Cast: Component-aligned 3d scene reconstruction from an rgb image},
  author={Yao, Kaixin and Zhang, Longwen and Yan, Xinhao and Zeng, Yan and Zhang, Qixuan and Xu, Lan and Yang, Wei and Gu, Jiayuan and Yu, Jingyi},
  journal={ACM Transactions on Graphics (TOG)},
  volume={44},
  number={4},
  pages={1--19},
  year={2025},
  publisher={ACM New York, NY, USA}
}

@article{ni2024phyrecon,
  title={Phyrecon: Physically plausible neural scene reconstruction},
  author={Ni, Junfeng and Chen, Yixin and Jing, Bohan and Jiang, Nan and Wang, Bin and Dai, Bo and Li, Puhao and Zhu, Yixin and Zhu, Song-Chun and Huang, Siyuan},
  journal={Advances in Neural Information Processing Systems},
  volume={37},
  pages={25747--25780},
  year={2024}
}

@inproceedings{agnew2021amodal,
  title={Amodal 3d reconstruction for robotic manipulation via stability and connectivity},
  author={Agnew, William and Xie, Christopher and Walsman, Aaron and Murad, Octavian and Wang, Yubo and Domingos, Pedro and Srinivasa, Siddhartha},
  booktitle={Conference on robot learning},
  pages={1498--1508},
  year={2021},
  organization={PMLR}
}

@inproceedings{bansal2023universal,
  title={Universal guidance for diffusion models},
  author={Bansal, Arpit and Chu, Hong-Min and Schwarzschild, Avi and Sengupta, Soumyadip and Goldblum, Micah and Geiping, Jonas and Goldstein, Tom},
  booktitle={Proceedings of the IEEE/CVF conference on computer vision and pattern recognition},
  pages={843--852},
  year={2023}
}

@inproceedings{yu2023freedom,
  title={Freedom: Training-free energy-guided conditional diffusion model},
  author={Yu, Jiwen and Wang, Yinhuai and Zhao, Chen and Ghanem, Bernard and Zhang, Jian},
  booktitle={Proceedings of the IEEE/CVF International Conference on Computer Vision},
  pages={23174--23184},
  year={2023}
}

@article{chung2022diffusion,
  title={Diffusion posterior sampling for general noisy inverse problems},
  author={Chung, Hyungjin and Kim, Jeongsol and Mccann, Michael T and Klasky, Marc L and Ye, Jong Chul},
  journal={The Eleventh International Conference on Learning Representations, ICLR 2023},
  year={2023}
}

@article{feng2025guidance,
  title={On the guidance of flow matching},
  author={Feng, Ruiqi and Yu, Chenglei and Deng, Wenhao and Hu, Peiyan and Wu, Tailin},
  journal={Forty-second International Conference on Machine Learning},
  year={2025}
}

@inproceedings{pan2023drag,
  title={Drag your gan: Interactive point-based manipulation on the generative image manifold},
  author={Pan, Xingang and Tewari, Ayush and Leimk{\"u}hler, Thomas and Liu, Lingjie and Meka, Abhimitra and Theobalt, Christian},
  booktitle={ACM SIGGRAPH 2023 conference proceedings},
  pages={1--11},
  year={2023}
}

@article{chen2024mvdrag3d,
  title={Mvdrag3d: Drag-based creative 3d editing via multi-view generation-reconstruction priors},
  author={Chen, Honghua and Lan, Yushi and Chen, Yongwei and Zhou, Yifan and Pan, Xingang},
  journal={arXiv preprint arXiv:2410.16272},
  year={2024}
}

@article{xu_instantmesh_2024,
  title={Instantmesh: Efficient 3d mesh generation from a single image with sparse-view large reconstruction models},
  author={Xu, Jiale and Cheng, Weihao and Gao, Yiming and Wang, Xintao and Gao, Shenghua and Shan, Ying},
  journal={arXiv preprint arXiv:2404.07191},
  year={2024}
}

@inproceedings{liu_zero-1--3_2023,
  title = {Zero-1-to-3: {Zero}-shot {One} {Image} to {3D} {Object}},
  author={Liu, Ruoshi and Wu, Rundi and Van Hoorick, Basile and Tokmakov, Pavel and Zakharov, Sergey and Vondrick, Carl},
  booktitle={Proceedings of the IEEE/CVF international conference on computer vision},
  pages={9298--9309},
  year={2023}
}

@inproceedings{mittal_autosdf_2022,
  title = {{AutoSDF}: {Shape} {Priors} for {3D} {Completion}, {Reconstruction} and {Generation}},
	 author={Mittal, Paritosh and Cheng, Yen-Chi and Singh, Maneesh and Tulsiani, Shubham},
  booktitle={Proceedings of the IEEE/CVF conference on computer vision and pattern recognition},
  pages={306--315},
  year={2022}
}

@inproceedings{cheng_sdfusion_2023,
  title = {{SDFusion}: {Multimodal} {3D} {Shape} {Completion}, {Reconstruction}, and {Generation}},
  author={Cheng, Yen-Chi and Lee, Hsin-Ying and Tulyakov, Sergey and Schwing, Alexander G and Gui, Liang-Yan},
  booktitle={Proceedings of the IEEE/CVF conference on computer vision and pattern recognition},
  pages={4456--4465},
  year={2023}
}

@inproceedings{chou_diffusion-sdf_2023,
  title = {Diffusion-{SDF}: {Conditional} {Generative} {Modeling} of {Signed} {Distance} {Functions}},
  author={Chou, Gene and Bahat, Yuval and Heide, Felix},
  booktitle={Proceedings of the IEEE/CVF international conference on computer vision},
  pages={2262--2272},
  year={2023}
}

@inproceedings{lin_diffsplat_2025,
  title = {{DiffSplat}: {Repurposing} {Image} {Diffusion} {Models} for {Scalable} {Gaussian} {Splat} {Generation}},
  author = {Lin, Chenguo and Pan, Panwang and Yang, Bangbang and Li, Zeming and Mu, Yadong},
  year = {2025},
  booktitle={The Thirteenth International Conference on Learning Representations}
}

@article{agarwal_scenecomplete_2024,
  title = {{SceneComplete}: {Open}-{World} {3D} {Scene} {Completion} in {Complex} {Real} {World} {Environments} for {Robot} {Manipulation}},
  author={Agarwal, Aditya and Singh, Gaurav and Sen, Bipasha and Lozano-P{\'e}rez, Tom{\'a}s and Kaelbling, Leslie Pack},
  journal={arXiv preprint arXiv:2410.23643},
  year={2024}
}

@inproceedings{ju_brushnet_2024,
  title = {{BrushNet}: {A} {Plug}-and-{Play} {Image} {Inpainting} {Model} with {Decomposed} {Dual}-{Branch} {Diffusion}},
  author={Ju, Xuan and Liu, Xian and Wang, Xintao and Bian, Yuxuan and Shan, Ying and Xu, Qiang},
  booktitle={European Conference on Computer Vision},
  pages={150--168},
  year={2024},
  organization={Springer}
}

@inproceedings{lugmayr_repaint_2022,
  title={RePaint: Inpainting using Denoising Diffusion Probabilistic Models}, 
  author={Lugmayr, Andreas and Danelljan, Martin and Romero, Andres and Yu, Fisher and Timofte, Radu and Van Gool, Luc},
  booktitle={Proceedings of the IEEE/CVF Conference on Computer Vision and Pattern Recognition},
  pages={11461--11471},
  year={2022}
}

@misc{zhao_hunyuan3d_2025,
  title = {{Hunyuan3D} 2.0: {Scaling} {Diffusion} {Models} for {High} {Resolution} {Textured} {3D} {Assets} {Generation}},
  author={Zhao, Zibo and Lai, Zeqiang and Lin, Qingxiang and Zhao, Yunfei and Liu, Haolin and Yang, Shuhui and Feng, Yifei and Yang, Mingxin and Zhang, Sheng and Yang, Xianghui and others},
  journal={arXiv preprint arXiv:2501.12202},
  year={2025}
}

@article{zhang_3dshape2vecset_2023,
  title = {{3DShape2VecSet}: {A} {3D} {Shape} {Representation} for {Neural} {Fields} and {Generative} {Diffusion} {Models}},
  author={Zhang, Biao and Tang, Jiapeng and Niessner, Matthias and Wonka, Peter},
  journal={ACM Transactions On Graphics (TOG)},
  volume={42},
  number={4},
  pages={1--16},
  year={2023},
  publisher={ACM New York, NY, USA}
}

@article{wu_amodal3r_2025,
  title = {{Amodal3R}: {Amodal} {3D} {Reconstruction} from {Occluded} {2D} {Images}},
  author={Wu, Tianhao and Zheng, Chuanxia and Guan, Frank and Vedaldi, Andrea and Cham, Tat-Jen},
  journal={arXiv preprint arXiv:2503.13439},
  year={2025}
}

@inproceedings{chen_using_nodate,
  title = {Using {Diffusion} {Priors} for {Video} {Amodal} {Segmentation}},
  author = {Chen, Kaihua and Ramanan, Deva and Khurana, Tarasha},
  booktitle={Proceedings of the Computer Vision and Pattern Recognition Conference},
  year = {2025},
  pages={22890--22900},
}

@inproceedings{ozguroglu_pix2gestalt_2024,
  title = {pix2gestalt: {Amodal} {Segmentation} by {Synthesizing} {Wholes}},
  booktitle={Proceedings of the IEEE/CVF conference on computer vision and pattern recognition},
  author = {Ozguroglu, Ege and Liu, Ruoshi and Surís, Dídac and Chen, Dian and Dave, Achal and Tokmakov, Pavel and Vondrick, Carl},
  year = {2024},
  pages = {3931--3940},
}

@inproceedings{xiang_structured_2025,
  title = {Structured {3D} {Latents} for {Scalable} and {Versatile} {3D} {Generation}},
  author={Xiang, Jianfeng and Lv, Zelong and Xu, Sicheng and Deng, Yu and Wang, Ruicheng and Zhang, Bowen and Chen, Dong and Tong, Xin and Yang, Jiaolong},
  booktitle={Proceedings of the Computer Vision and Pattern Recognition Conference},
  pages={21469--21480},
  year={2025}
}

@inproceedings{xie2022smartbrushtextshapeguided,
  title={Smartbrush: Text and shape guided object inpainting with diffusion model},
  author={Xie, Shaoan and Zhang, Zhifei and Lin, Zhe and Hinz, Tobias and Zhang, Kun},
  booktitle={Proceedings of the IEEE/CVF conference on computer vision and pattern recognition},
  pages={22428--22437},
  year={2023}
}

@inproceedings{zhang2023adding,
  title={Adding Conditional Control to Text-to-Image Diffusion Models}, 
  author={Zhang, Lvmin and Rao, Anyi and Agrawala, Maneesh},
  booktitle={Proceedings of the IEEE/CVF international conference on computer vision},
  pages={3836--3847},
  year={2023}
}

@inproceedings{bianchini2025vysics,
  title={Vysics: Object Reconstruction Under Occlusion by Fusing Vision and Contact-Rich Physics},
  author={Bianchini, Bibit and Zhu, Minghan and Sun, Mengti and Jiang, Bowen and Taylor, Camillo J and Posa, Michael},
  booktitle={Robotics: Science and Systems (RSS)},
  year={2025}
}

@article{mou2023dragondiffusion,
  title={Dragondiffusion: Enabling drag-style manipulation on diffusion models},
  author={Mou, Chong and Wang, Xintao and Song, Jiechong and Shan, Ying and Zhang, Jian},
  journal={arXiv preprint arXiv:2307.02421},
  year={2023}
}

@inproceedings{shi2024dragdiffusionharnessingdiffusionmodels,
  title={DragDiffusion: Harnessing Diffusion Models for Interactive Point-based Image Editing}, 
  author={Shi, Yujun and Xue, Chuhui and Liew, Jun Hao and Pan, Jiachun and Yan, Hanshu and Zhang, Wenqing and Tan, Vincent YF and Bai, Song},
  booktitle={Proceedings of the IEEE/CVF Conference on Computer Vision and Pattern Recognition},
  pages={8839--8849},
  year={2024}
}

@inproceedings{wen2023bundlesdfneural6doftracking,
  title={BundleSDF: Neural 6-DoF Tracking and 3D Reconstruction of Unknown Objects}, 
  author={Wen, Bowen and Tremblay, Jonathan and Blukis, Valts and Tyree, Stephen and M{\"u}ller, Thomas and Evans, Alex and Fox, Dieter and Kautz, Jan and Birchfield, Stan},
  booktitle={Proceedings of the IEEE/CVF Conference on Computer Vision and Pattern Recognition},
  pages={606--617},
  year={2023}
}

@inproceedings{wen2024foundationpose,
  title={FoundationPose: Unified 6d pose estimation and tracking of novel objects},
  author={Wen, Bowen and Yang, Wei and Kautz, Jan and Birchfield, Stan},
  booktitle={Proceedings of the IEEE/CVF Conference on Computer Vision and Pattern Recognition},
  pages={17868--17879},
  year={2024}
}

@inproceedings{rombach2022highresolutionimagesynthesislatent,
  title={High-Resolution Image Synthesis with Latent Diffusion Models}, 
  author={Rombach, Robin and Blattmann, Andreas and Lorenz, Dominik and Esser, Patrick and Ommer, Bj{\"o}rn},
  booktitle={Proceedings of the IEEE/CVF Conference on Computer Vision and Pattern Recognition},
  pages={10684--10695},
  year={2022}
}

@inproceedings{szymanowicz2023viewsetdiffusion0imageconditioned3d,
  title={Viewset Diffusion: (0-)Image-Conditioned 3D Generative Models from 2D Data}, 
  author={Szymanowicz, Stanislaw and Rupprecht, Christian and Vedaldi, Andrea},
  booktitle={Proceedings of the IEEE/CVF international conference on computer vision},
  pages={8863--8873},
  year={2023}
}

@inproceedings{xu2023dmv3ddenoisingmultiviewdiffusion,
  title={DMV3D: Denoising Multi-View Diffusion using 3D Large Reconstruction Model}, 
  author={Xu, Yinghao and Tan, Hao and Luan, Fujun and Bi, Sai and Wang, Peng and Li, Jiahao and Shi, Zifan and Sunkavalli, Kalyan and Wetzstein, Gordon and Xu, Zexiang and others},
  booktitle={The Twelfth International Conference on Learning Representations},
  year={2024}
}

@inproceedings{poole2022dreamfusiontextto3dusing2d,
  title={DreamFusion: Text-to-3D using 2D Diffusion}, 
  author={Poole, Ben and Jain, Ajay and Barron, Jonathan T and Mildenhall, Ben},
  booktitle={The Eleventh International Conference on Learning Representations},
  year={2023},
}

@article{wang2023imagedreamimagepromptmultiviewdiffusion,
  title={ImageDream: Image-Prompt Multi-view Diffusion for 3D Generation}, 
  author={Wang, Peng and Shi, Yichun},
  journal={arXiv preprint arXiv:2312.02201},
  year={2023}
}

@inproceedings{shi2024mvdreammultiviewdiffusion3d,
  title={MVDream: Multi-view Diffusion for 3D Generation}, 
  author={Shi, Yichun and Wang, Peng and Ye, Jianglong and Long, Mai and Li, Kejie and Yang, Xiao},
  year={2024},
  booktitle={The Twelfth International Conference on Learning Representations}
}

@article{liu2023one2345singleimage3d,
  title={One-2-3-45: Any Single Image to 3D Mesh in 45 Seconds without Per-Shape Optimization}, 
  author={Liu, Minghua and Xu, Chao and Jin, Haian and Chen, Linghao and Varma T, Mukund and Xu, Zexiang and Su, Hao},
  journal={Advances in Neural Information Processing Systems},
  volume={36},
  pages={22226--22246},
  year={2023}
}

@inproceedings{li2023instant3dfasttextto3dsparseview,
  title={Instant3D: Fast Text-to-3D with Sparse-View Generation and Large Reconstruction Model}, 
  author={Li, Jiahao and Tan, Hao and Zhang, Kai and Xu, Zexiang and Luan, Fujun and Xu, Yinghao and Hong, Yicong and Sunkavalli, Kalyan and Shakhnarovich, Greg and Bi, Sai},
  year={2024},
  booktitle={The Twelfth International Conference on Learning Representations}
}

@inproceedings{long2023wonder3dsingleimage3d,
  title={Wonder3D: Single Image to 3D using Cross-Domain Diffusion}, 
  author={Long, Xiaoxiao and Guo, Yuan-Chen and Lin, Cheng and Liu, Yuan and Dou, Zhiyang and Liu, Lingjie and Ma, Yuexin and Zhang, Song-Hai and Habermann, Marc and Theobalt, Christian and others},
  booktitle={Proceedings of the IEEE/CVF conference on computer vision and pattern recognition},
  pages={9970--9980},
  year={2024}
}

@article{chang2015shapenetinformationrich3dmodel,
  title={ShapeNet: An Information-Rich 3D Model Repository}, 
  author={Chang, Angel X and Funkhouser, Thomas and Guibas, Leonidas and Hanrahan, Pat and Huang, Qixing and Li, Zimo and Savarese, Silvio and Savva, Manolis and Song, Shuran and Su, Hao and others},
  journal={arXiv preprint arXiv:1512.03012},
  year={2015}
}

@inproceedings{chen20253dtopiaxlscalinghighquality3d,
  title={3DTopia-XL: Scaling High-quality 3D Asset Generation via Primitive Diffusion}, 
  author={Chen, Zhaoxi and Tang, Jiaxiang and Dong, Yuhao and Cao, Ziang and Hong, Fangzhou and Lan, Yushi and Wang, Tengfei and Xie, Haozhe and Wu, Tong and Saito, Shunsuke and others},
  booktitle={Proceedings of the Computer Vision and Pattern Recognition Conference},
  pages={26576--26586},
  year={2025}
}

@inproceedings{lan2025gaussiananythinginteractivepointcloud,
  title={GaussianAnything: Interactive Point Cloud Flow Matching For 3D Object Generation}, 
  author={Yushi, LAN and Zhou, Shangchen and Lyu, Zhaoyang and Hong, Fangzhou and Yang, Shuai and Dai, Bo and Pan, Xingang and Loy, Chen Change},
  booktitle={The Thirteenth International Conference on Learning Representations},
  year={2025}
}

@article{zhang2024claycontrollablelargescalegenerative,
  title={CLAY: A Controllable Large-scale Generative Model for Creating High-quality 3D Assets}, 
  author={Zhang, Longwen and Wang, Ziyu and Zhang, Qixuan and Qiu, Qiwei and Pang, Anqi and Jiang, Haoran and Yang, Wei and Xu, Lan and Yu, Jingyi},
  journal={ACM Transactions on Graphics (TOG)},
  volume={43},
  number={4},
  pages={1--20},
  year={2024},
  publisher={ACM New York, NY, USA}
}

@inproceedings{deitke2023objaverse,
  title={Objaverse: A universe of annotated 3d objects},
  author={Deitke, Matt and Schwenk, Dustin and Salvador, Jordi and Weihs, Luca and Michel, Oscar and VanderBilt, Eli and Schmidt, Ludwig and Ehsani, Kiana and Kembhavi, Aniruddha and Farhadi, Ali},
  booktitle={Proceedings of the IEEE/CVF conference on computer vision and pattern recognition},
  pages={13142--13153},
  year={2023}
}

@inproceedings{yuan2019pcnpointcompletionnetwork,
  title={PCN: Point Completion Network}, 
  author={Yuan, Wentao and Khot, Tejas and Held, David and Mertz, Christoph and Hebert, Martial},
  booktitle={2018 international conference on 3D vision (3DV)},
  pages={728--737},
  year={2018},
  organization={IEEE}
}

@inproceedings{yu2021pointrdiversepointcloud,
  title={PoinTr: Diverse Point Cloud Completion with Geometry-Aware Transformers}, 
  author={Yu, Xumin and Rao, Yongming and Wang, Ziyi and Liu, Zuyan and Lu, Jiwen and Zhou, Jie},
  booktitle={Proceedings of the IEEE/CVF international conference on computer vision},
  pages={12498--12507},
  year={2021}
}

@inproceedings{chen2023anchorformer,
  title={Anchorformer: Point cloud completion from discriminative nodes},
  author={Chen, Zhikai and Long, Fuchen and Qiu, Zhaofan and Yao, Ting and Zhou, Wengang and Luo, Jiebo and Mei, Tao},
  booktitle={Proceedings of the IEEE/CVF conference on computer vision and pattern recognition},
  pages={13581--13590},
  year={2023}
}

@inproceedings{mou2023t2iadapterlearningadaptersdig,
  title={T2I-Adapter: Learning Adapters to Dig out More Controllable Ability for Text-to-Image Diffusion Models}, 
  author={Mou, Chong and Wang, Xintao and Xie, Liangbin and Wu, Yanze and Zhang, Jian and Qi, Zhongang and Shan, Ying},
  booktitle={Proceedings of the AAAI conference on artificial intelligence},
  volume={38},
  number={5},
  pages={4296--4304},
  year={2024}
}

@article{yao2025castcomponentaligned3dscene,
  title={CAST: Component-Aligned 3D Scene Reconstruction from an RGB Image}, 
  author={Yao, Kaixin and Zhang, Longwen and Yan, Xinhao and Zeng, Yan and Zhang, Qixuan and Xu, Lan and Yang, Wei and Gu, Jiayuan and Yu, Jingyi},
  journal={ACM Transactions on Graphics (TOG)},
  volume={44},
  number={4},
  pages={1--19},
  year={2025},
  publisher={ACM New York, NY, USA}
}

@article{oquab2023dinov2,
  title={Dinov2: Learning robust visual features without supervision},
  author={Oquab, Maxime and Darcet, Timoth{\'e}e and Moutakanni, Th{\'e}o and Vo, Huy and Szafraniec, Marc and Khalidov, Vasil and Fernandez, Pierre and Haziza, Daniel and Massa, Francisco and El-Nouby, Alaaeldin and others},
  journal={Transactions on Machine Learning Research Journal},
  year={2024}
}

@inproceedings{downs2022googlescannedobjectshighquality,
  title={Google Scanned Objects: A High-Quality Dataset of 3D Scanned Household Items}, 
  author={Downs, Laura and Francis, Anthony and Koenig, Nate and Kinman, Brandon and Hickman, Ryan and Reymann, Krista and McHugh, Thomas B and Vanhoucke, Vincent},
  booktitle={2022 International Conference on Robotics and Automation (ICRA)},
  pages={2553--2560},
  year={2022},
  organization={IEEE}
}

@article{Shen_2023,
  title={Flexible Isosurface Extraction for Gradient-Based Mesh Optimization},
  author={Shen, Tianchang and Munkberg, Jacob and Hasselgren, Jon and Yin, Kangxue and Wang, Zian and Chen, Wenzheng and Gojcic, Zan and Fidler, Sanja and Sharp, Nicholas and Gao, Jun},
  journal={ACM Transactions on Graphics (TOG)},
  volume={42},
  number={4},
  pages={1--16},
  year={2023},
  publisher={ACM New York, NY, USA}
}

\end{document}